\documentclass[letterpaper, 10 pt, conference]{ieeeconf}  

\IEEEoverridecommandlockouts                              

\usepackage{comment}

\title{\LARGE \bf
 Robot-Enabled Machine Learning-Based Diagnosis of Gastric Cancer Polyps Using 
  Partial Surface Tactile Imaging
}

\author{Siddhartha Kapuria*$^{1}$, Jeff Bonyun*$^{1}$, Yash Kulkarni$^{1}$, Naruhiko Ikoma$^{2}$, Sandeep Chinchali$^{1}$, \\ and Farshid Alambeigi$^{1}$
\thanks{*Authors had equal contribution to this work}
\thanks{This research was supported by the National
Cancer Institute of the National Institutes of Health under Award Number
R21CA280747}
\thanks{$^{1}$Siddhartha Kapuria, Jeff Bonyun, Yash Kulkarni, Sandeep Chinchali, and Farshid Alambeigi are with Texas Robotics, University of Texas at Austin, Austin, TX, USA. email: {\tt\small \{skapuria, jbonyun, kulkarni.yash08, sandeepc\}@utexas.edu, and farshid.alambeigi@austin.utexas.edu}}%
\thanks{$^{2}$Naruhiko Ikoma is with the Department of Surgical Oncology, Division of Surgery, The University of Texas MD Anderson Cancer Center, Houston, TX, USA, 77030. email: {\tt\small nikoma@mdanderson.org}}
}
\let\labelindent\relax
\usepackage{enumitem}
\usepackage{graphicx}
\graphicspath{ {./Figures/} }
\usepackage{subcaption}
\usepackage{multirow}
\usepackage{hyperref}
\hypersetup{
    colorlinks=true,
    linkcolor=blue,
    filecolor=magenta,      
    urlcolor=cyan,
    }
\usepackage{makecell}
\usepackage{booktabs}
\usepackage{fancyhdr}
\fancypagestyle{firstpage}{%
  \lhead{Accepted for publication at the 2024 IEEE/RSJ International Conference on Intelligent Robots and Systems (IROS 2024)}
}

\usepackage{color}

\begin{document}

\maketitle
\thispagestyle{firstpage}

\begin{abstract}
In this paper, to collectively address the existing limitations on endoscopic diagnosis of Advanced Gastric Cancer (AGC) Tumors, for the first time, we propose (i) utilization and evaluation of  our recently developed Vision-based Tactile Sensor (VTS), and  (ii) a complementary Machine Learning (ML) algorithm for  classifying  tumors using their textural features. Leveraging a seven DoF robotic manipulator and unique custom-designed and additively-manufactured realistic AGC tumor phantoms, we demonstrated the advantages of automated  data collection using the VTS addressing the problem of data scarcity and biases encountered in traditional ML-based approaches.  Our synthetic-data-trained ML model was successfully evaluated and compared with traditional ML models utilizing various statistical metrics even under mixed morphological characteristics and partial sensor contact. 

\end{abstract}

\begin{figure}[t!]
		\centering
		\includegraphics[width=0.85\linewidth]{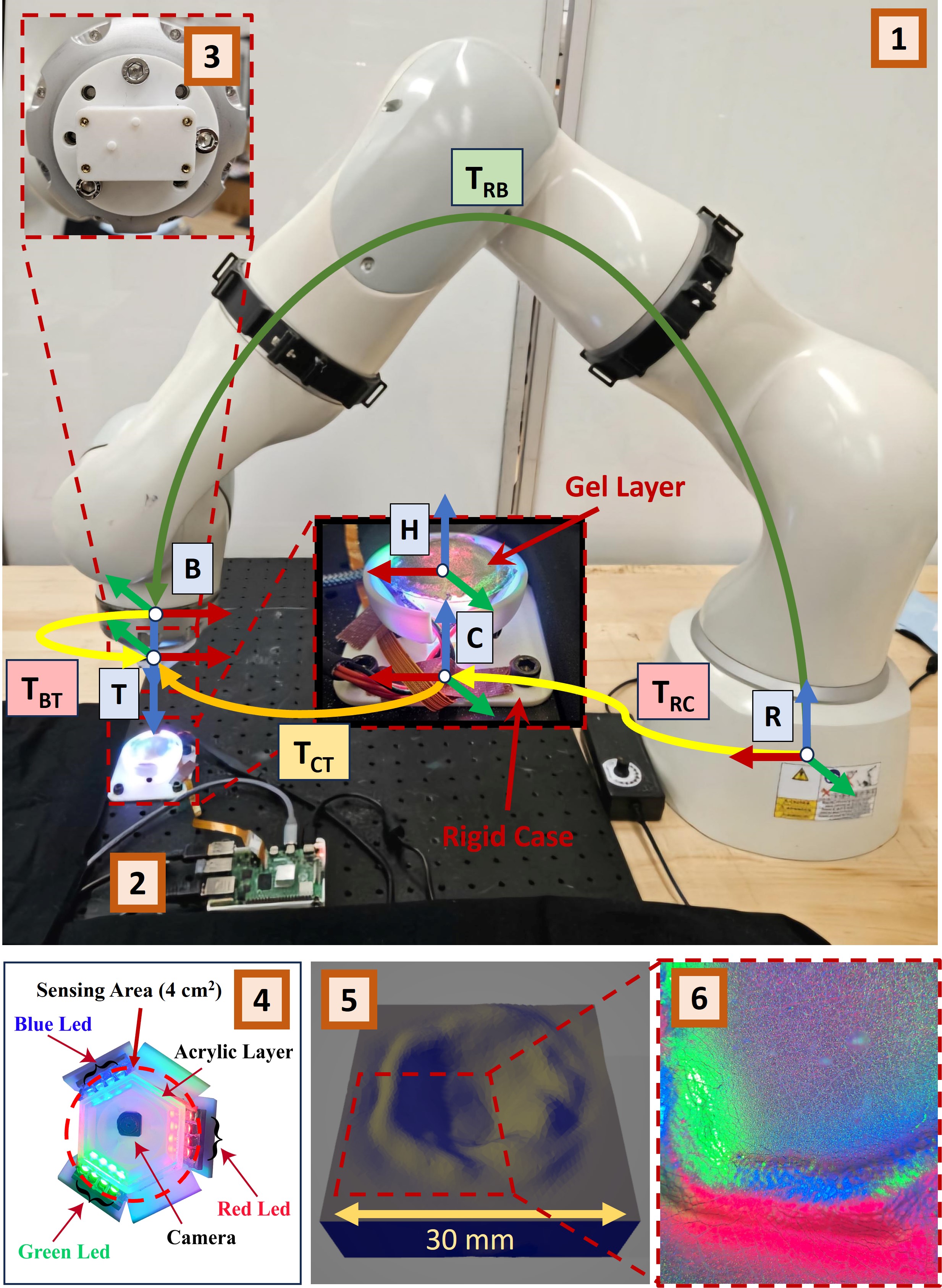}
		\vspace*{-1mm}
\caption{Experimental Setup including: (1) KUKA LBR
Med 14 R820 (KUKA AG); (2) Raspberry Pi 4 Model B; (3) 3D printed mounting plate for the tumor phantoms; (4) Top view of HySenSe sensor showing all components; (5) Example CAD model of synthetic AGC polyp phantom; (6) Example partial textural image output of AGC tumor phantom. Figure also shows the defined reference frames \textit{R}: Robot/World, \textit{B}: Robot Flange, \textit{C}: Camera, \textit{T}: Target, and \textit{H}: HySenSe base.}
\label{fig:setup}
    \vspace*{-8mm}
\end{figure}

\section{INTRODUCTION}

\begin{figure}[t!]
		\centering
  		\vspace*{2mm}
		\includegraphics[width=0.49\textwidth]{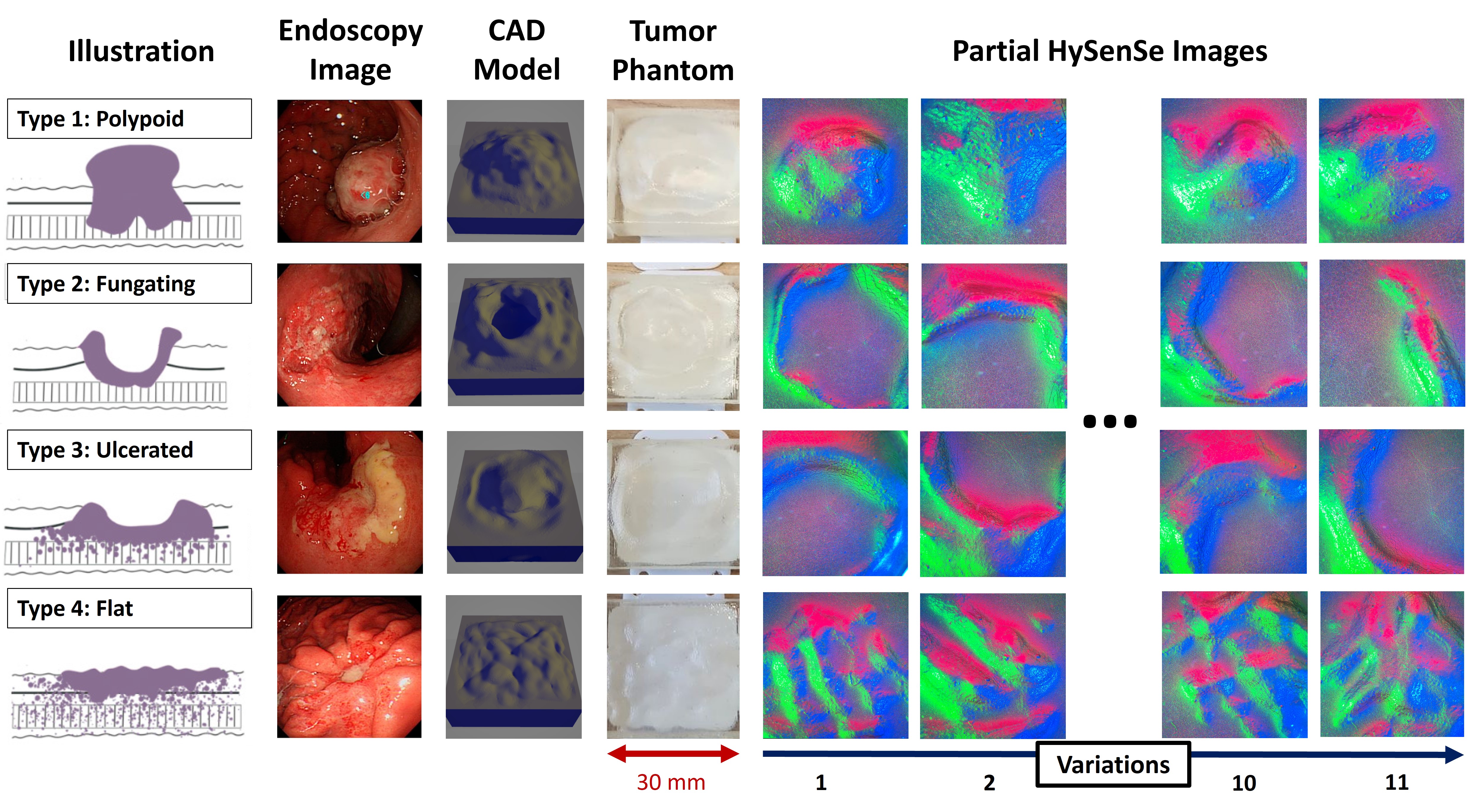}
		\vspace*{-5mm}
\caption{The first column shows the schematics  \cite{Wang2015ComparingSO} of 4 types of AGC tumors under the Borrmann classification. The second column indicates the corresponding real clinical endoscopic images \cite{Hosoda2018ReemergingRO}. The third and fourth columns show a sample designed CAD model and 3D-printed tumor for each class. Other columns show corresponding VTS  outputs. }
\label{fig:tumors}
    \vspace*{-2.1em}
\end{figure}

Gastric cancer (GC) is  the fifth most commonly diagnosed cancer worldwide and the fourth leading cause of cancer-related mortality \cite{Sung2021GlobalCS}. A major contributor to this challenge is the fact that a substantial portion — up to 62\% — of GC cases are detected at advanced stages, contributing to poorer overall survival rates compared to cases identified at early stages \cite{seer2021Cancer}. Upper endoscopy is the primary method for the initial detection of GC lesions as it allows for an inside view of the gastric tract lining where tumors originate. At the advanced GC  (AGC) stages, tumors  have infiltrated the muscularis propria  \cite{Gore2013StomachMT} and can be identified and classified through their morphological characteristics (i.e., their geometry and texture) visible through the images provided by an endoscope.  Borrmann classification \cite{Gore2013StomachMT}   is a common approach used by clinicians to morphologically classify GC polyps into four types of  polypoid (Type 1), fungating (Type 2), ulcerated (Type 3), and infiltrating or Flat (Type 4) (see Fig. \ref{fig:tumors}). Nevertheless,  inter-class variance of each type of polyps and solely relying on morphology of the GC polyps in Borrmann classification has resulted in a high-degree of disagreement and inconsistency in decision-making among clinicians \cite{Lockhart2009EpidemiologyOG}. Therefore, long-term specific training and experience is needed to detect GC properly using endoscopic images and Borrmann classification
\cite{Zhang2015TrainingIE}. Furthermore, similar to many vision-based endoscopic diagnoses (e.g., colonoscopy and laparoscopy), the limited resolution of endoscopic video cameras, visual occlusions, lack of sufficient steerability of the endoscopic devices, and lighting changes within the body make reliable diagnosis of GC polyps even more challenging \cite{Ginzburg2007ComplicationsOE}.

To address the above-mentioned limitations, Artificial Intelligence (AI) methods utilizing Machine Learning (ML) have been employed in different modalities, such as histopathological images or endoscopic videos, to detect and classify GC tumors.  For example, Li et al. \cite{Li2018DeepLB} used a custom Deep Learning (DL) based framework for automatic cancer identification from histopathological images. Using the same modality, Huang et al. \cite{Huang2021AccurateDA} developed an in-house DL approach - GastroMIL - for differentiating between cancerous and healthy tissue. For endoscopic videos, Hirasawa et al. \cite{Hirasawa2018ApplicationOA} used a CNN-based Single Shot MultiBox Detector to successfully detect cancerous lesions. Taking a step further to address the limitations of traditional endoscopy, Xia et al. \cite{Xia2020UseOA} used a magnetically controlled capsule endoscope coupled with an ML model for \textit{in-vivo} classification of GC.

A common limitation of such ML approaches is the limited availability and access to the large, balanced datasets \cite{Fotouhi2019ACD}. This has also been recognized by The American Medical Association, which passed policy recommendations in 2018 for identifying and mitigating bias in data during the testing or deployment of AI/ML-based software to prevent introducing or exacerbating healthcare disparities \cite{ama2023}. DL approaches especially require large amounts of data to be able to generalize over unknown inputs. This is particularly important in the medical domain, where there are several hurdles to obtaining patient data such as time dependence, availability of relevant clinical cases, and privacy concerns \cite{Christou2021ChallengesAO}. Limited data can lead to (1) spectrum bias and (2) overfitting  
\cite{Christou2021ChallengesAO}. 
Furthermore, in the case of AGC tumors, there is a very high degree of inter-class variance in the morphological characteristics, which can vary considerably amongst patients, hindering the preparation of a rich, well-balanced dataset \cite{Lockhart2009EpidemiologyOG}. This issue of limited data can be partially mitigated through transfer learning, which has now become a staple of modern DL frameworks \cite{Jin2020ArtificialII}. However, the limitation of class imbalance persists through this technique. This is important to consider since even if datasets mirror the real-world distribution, rarer cases may not have adequate representation. In such scenarios, where rare cancer cases are both inherently infrequent and underrepresented in the data, ML models may struggle to learn the distinct features associated with these cases.

To address the aforementioned limitations of existing endoscopic procedures, recently we have developed a novel Vision-based Tactile Sensor (VTS) (shown in Fig. \ref{fig:setup}) and introduced the surface tactile imaging modality   for early diagnosis of colorectal cancer (CRC) polyps \cite{Kara2023ASH, Venkatayogi2022PitPatternCO, Kara2023abme}. As opposed to normal endoscopic images, VTS provides high-resolution $\sim$50 $\mu m$ textural images of the polyps that can potentially improve their classification. However, the size of CRC polyps  are  smaller than AGC tumors (i.e., $\sim$1 cm$\times$1 cm versus $\sim$4 cm$\times$4 cm) \cite{Kudo1994ColorectalTA,Gore2013StomachMT}. As shown in Fig. \ref{fig:setup}, since the sensing area of the VTS is limited, only partial textural images of AGC tumors can be captured using VTS, making their data collection and  classification very challenging.  

To collectively address these limitations and towards developing an ML-based diagnostic assistance for AGC  using our VTS, in this work and as our main \textit{contributions}, we propose (i) to use and evaluate, for the first time, the potentials of utilizing the VTS in classifying AGC tumors using their textural features; (ii) a complementary ML-based diagnostic tool that can leverage this new modality to sensitively classify AGC lesions; and (iii) a robot-assisted data collection procedure to ensure the ML model is trained on a large and balanced dataset. We train the ML models on partial textural data semi-autonomously collected from our unique 3D-printed AGC tumor phantoms. We also use appropriate statistical metrics during evaluation to show that the proposed ML models can reliably and sensitively classify the AGC lesions even under mixed morphological conditions and partial tumor coverage.

\section{MATERIALS AND METHODS}

\subsection{Vision Based Tactile Sensor (VTS)}
\label{sec:hysense}
In this study, we employed our recently developed VTS called HySenSe, as outlined in~\cite{Kara2022HySenSeAH}, to acquire high-fidelity textural images of AGC tumor phantoms. As depicted in Fig. \ref{fig:setup}, the HySenSe sensor comprises: (I) a flexible silicone membrane interacting directly with polyp phantoms, (II) an optical module (Arducam 1/4 inch 5 MP camera) capturing minute deformations in the gel layer during interactions with a polyp phantom, (III) a transparent acrylic plate offering support to the gel layer, (IV) an array of Red, Green, and Blue LEDs for internal illumination aiding depth perception, and (V) a sturdy frame supporting the entire structure. 
Operating on the principle that the deformations resulting from the interaction between the deformable membrane and the surface of AGC tumors can be visually captured, the HySenSe sensor provides high-fidelity textural images, demonstrating proficiency across various tumor characteristics such as surface texture, hardness, type, and size \cite{Kara2023abme}. This capability is maintained even at extremely low interaction forces, as detailed in \cite{Kara2022HySenSeAH}. Additionally, due to the arrangement of the LEDs within the sensor, different deformations have different lighting. This means that if the interaction force is fixed, the textural images implicitly encode the stiffness characteristics of the surface in contact as well. Due to these advantages, in our previous works \cite{Kara2023ASH, Venkatayogi2022PitPatternCO, Kara2023abme}, we have successfully used this sensor to differentiate between classes of CRC polyps, which are also distinguished by their morphological characteristics, namely the surface texture presented and the polyp stiffness. However, at this point, it must be noted that CRC polyps and AGC tumors differ greatly in size (i.e., $\sim$1 cm$\times$1 cm versus $\sim$4 cm$\times$4 cm). Since building a larger HySenSe sensor is impractical in a clinical setting, we are limited by the area we cover using the sensor. In its current form, the area coverage is around 4 cm$^{2}$ (see Fig. \ref{fig:setup}.4). Therefore it is impossible to entirely cover an average AGC tumor. To address this issue and cover sufficient amount of data, we opted to employ a semi-autonomous robotic system, to sequentially cover the whole AGC surface and collect data discussed in detail in Section \ref{sec:setup}.

\vspace*{-0.2em}
\subsection{Realistic Tumor Phantoms}
Towards addressing the limited availability of large, balanced clinical datasets in medical imaging and given this new imaging modality, we opted to meticulously design and manufacture realistic approximations of AGC tumors (see Fig. \ref{fig:tumors}). To design the phantoms, and without loss of generality evaluate the performance of the utilized VTS, we followed 4 different types of Borrmann's classification system. The four types specifically are Type I: fungating type; Type II: carcinomatous ulcer without infiltration of the surrounding mucosa; Type III: carcinomatous ulcer with infiltration of the surrounding mucosa; Type IV: a diffuse infiltrating carcinoma (linitis plastic) \cite{Gore2013StomachMT}. It is known that the stiffness of the affected area is more than that of the surrounding regions, which makes it easier for clinicians to differentiate between tumorous sections and healthy tissue \cite{Gore2013StomachMT}. However, this classification has a considerable overlap (especially between Type II and Type III) due to the mixed morphological characteristics of these lesions, making a manual diagnosis through observation difficult \cite{Lockhart2009EpidemiologyOG}.

Fig. \ref{fig:tumors} illustrates a few representative fabricated AGC lesion phantoms and their dimensions. To avoid the issue of data imbalance, in contrast with real patient datasets, we opted to have each tumor class equally represented in the dataset by designing 11 variations of each class (total 44 polyps). As shown in Fig. \ref{fig:tumors}, based on the realistic AGC polyps, the designs first were conceptualized in Blender Software (The Blender Foundation) to make use of the free-form sculpting tool, then imported into Solidworks (Dassault Systèmes)  in order to demarcate the different regions with varying stiffness (tumor versus healthy tissue). The high-resolution, realistic lesion phantoms were manufactured using a Digital Anatomy Printer (J750, Stratasys, Ltd) and materials with diverse properties: (M1) Tissue Matrix/Agilus DM 400, (M2) a mixture of Tissue Matrix and Agilus 30 Clear, and (M3) Vero PureWhite. Hardness measurements were obtained using a Shore 00 scale durometer (Model 1600 Dial Shore 00, Rex Gauge Company). M1 has Shore hardness A 1-2, M2 has A 30-40, and M3 has D 83-86. These differing material properties allowed us to make the tumor sections (using M2) stiffer than the surrounding healthy tissue (using M1). M3 was used to print the supporting rigid backplate to be mounted onto the robot flange. Each tumor was printed across a working area of 3 cm $\times$ 3 cm, which represents the lower end of AGC tumor sizes.

\subsection{Experimental Setup and Robotic Data Collection}
\label{sec:setup}
In our previous works utilizing HySenSe for CRC polyp classification (see \cite{Kara2023ASH}, \cite{Venkatayogi2022PitPatternCO}, \cite{Kara2023abme}), the high-fidelity textural images were captured manually using a setup including a force gauge mounted on a linear stage. This manual and tedious procedure limited our data collection capabilities, which was not efficient for this study. Furthermore, as discussed in Section \ref{sec:hysense}, AGC tumors are much larger than CRC polyps, making it impossible for the VTS to completely capture the whole textural area of the  tumor phantoms. To overcome this limitation, in this work, we used a robotic manipulator to automate the image collection procedure. Using the robotic arm shown in Fig. \ref{fig:setup}, many different variations of partial contact from different angles were captured to form our dataset, which allowed the trained ML model to be more generalized. 

The experimental setup for data collection is illustrated in Fig \ref{fig:setup} and consists of the following: \textbf{(1) Robot Manipulator:} We used a KUKA LBR Med 14 R820 (KUKA AG) which has seven degrees of freedom (DoF), a large operating envelope, and integrated force sensing. We used ROS as a bridge, with the \verb|iiwa_stack| project presented in \cite{hennersperger2017towards}, to provide high-level control of the onboard Java environment. \textbf{(2) Workspace: }The arm was rigidly attached to a work table. An optical table allows consistent positioning of items in the robot’s coordinate frame. \textbf{(3) HySenSe: } The sensor was manufactured by our lab, using the methodology provided in \cite{Kara2022HySenSeAH}, and attached to the optical table.  Images are captured by the camera, a 5MP Arducam with 1/4” sensor (model OV5647, Arducam). \textbf{(4) Raspberry Pi: } The camera is controlled by a Raspberry Pi 4B over the Arducam's camera ribbon cable. The Raspberry Pi ran Python software that continuously listened for an external ROS message trigger, which caused it to capture a 2592 $\times$ 1944 image and publish it to ROS. \textbf{(5) Sample Mount: } An adapting mount was 3D printed in PLA. One side attached to the robot flange. The other side offered two locating pins and four M2 screw holes, which ensure repeatable position and orientation of samples \textbf{(6) Polyp Phantom Samples: } The polyp samples were constructed as described above. Each was attached, in turn, to the sample mount using M2 screws. \textbf{(7) Command and Control: } An Ubuntu 20.04 system was used to control the ROS components. This computer ran the \verb|roscore| master and the high-level Python command scripts.

To collect textural images with HySenSe, the robotic manipulator was commanded via ROS to position the phantom of interest into random positions with different angles of contact, while maintaining the interaction force under a threshold of 3N using the arm's internal force sensor. As opposed to our previous manula procedures performed in \cite{Kara2023abme}, the only manual step involved was installing each target AGC phantom onto the sample mount on the robot end effector, and the remainder of the process was automated in software, reducing the time and workload required.

\subsection{Calibration and Registration Procedures}
While the apparatus was assembled carefully, there was no way to ensure precise positioning to the sub-millimeter level. This level of accuracy was necessary to successfully automate the action of pressing down an AGC tumor phantom onto the HySenSe gel with a measured and limited force of interaction.
 Thus, a registration step was performed to find the unknown transformation matrices amongst the different components. We divided our test setup into five frames of reference, visualized in Fig. \ref{fig:setup}, with the objective of determining the transformation matrices between these frames. Some of these transformations were known. The transformation from robot base ($R$) to robot flange ($B$), $T_{RB}$, is provided through the KUKA software. The transformation from the camera frame ($C$) to a known target ($T$), $T_{CT}$, was calculated by using a standard checkerboard calibration method. The intrinsic matrix follows the pinhole camera model described in \cite{Hartley_2000} and the 10-parameter distortion model follows \cite{Brown_1966} and \cite{Zhang_2000}. 
 The remaining transformations were unknown, including the ones from the robot base to the camera frame ($T_{RC}$), from the robot’s flange to the tumor phantom ($T_{BT}$), and from the camera frame to the plastic plate that holds the HySenSe gel  ($T_{CH}$).

\begin{table}[t]
\vspace{0.5em}
\centering
\caption{Hyperparameter Space}
\begin{tabular}{@{}lc@{}}
\toprule
\textbf{Hyperparameter}    & \textbf{Range/Options}                                        \\ \midrule
Learning Rate               & (min = 0.001, max = 0.1)                                      \\ \midrule
Learning Rate Scheduler    & \makecell{Step \\ Reduce on Plateau \\ 1cycle\\ Cosine Annealing} \\ \midrule
Optimizer                  & \makecell{Stochastic Gradient Descent \\ Adam \\ Adabound}        \\ \midrule
Weight Decay               & (min = 0, max = 0.1)                                          \\ \bottomrule
\end{tabular}
\label{tab:hparams}
\vspace{-2.5em}
\end{table}

Using the checkerboard calibration images as data, we performed $AX=XB$ calibration using the separable method \cite{Horaud_1995}. This provided us with estimates of $T_{RC}$ and $T_{BT}$. The resulting registration was still noisy, and fine-tuning was done with a simple grid pattern, adjusting the robot position until it was centered and square in the camera image.
The final part of the registration was finding $T_{CH}$ by bringing a flat sample plate down until it was within $1mm$ of the HySenSe backing plate, with the gel removed, and checking that it was parallel with a feeler gauge.

\subsection{Dataset and Pre-processing} 
This semi-automated data collection setup enabled us to collect 50 variations of orientation and contact of the AGC tumor phantoms with the HySenSe in 44 experiments (one for each polyp), leading to a total of 2200 images in the textural image dataset, with 550 unique images in each class. 
This dataset was then split into training and test sets while ensuring that the split was performed at a tumor level and not the image level. In other words, all textural images belonging to one tumor were kept within the same split. This was done in order to ensure the model was not being evaluated on partially seen data. This resulted in 1600 images from 32 unique tumors in the training set and 600 images from 12 unique tumors in the test set. The output from HySenSe, with original dimensions of $2592 \times 1944$ pixels, was resized to a uniform size of $224 \times 224$ pixels, ensuring consistent input dimensions for all models. To enhance the algorithm's generalization capability, a range of geometric transformations, including random cropping, vertical and horizontal flips, and random rotations spanning from -45° to 45°, were incorporated as augmentations.  Further to enhance generalizability of data augmentation, we also included Gaussian blur and Gaussian noise with strengths $\sigma$ ranging from 1 to 256 for blur, and $\sigma$ from 1 to 50 for noise \cite{Kapuria2023TowardsRC}. The maximums for these values were chosen to represent a level well beyond realistic worst-case scenarios \cite{Kapuria2023TowardsRC}. Each augmentation had an independent occurrence probability of 0.5, contributing to the algorithm's overall robustness. The images in the holdout test dataset, comprising 600 samples, were only resized to fit as inputs to all models.

\subsection{Deep Learning Model}

In this study, we opted to use the Dilated ResNet architecture for AGC tumor classification. The architecture's effectiveness stems from its capacity to mitigate issues like exploding gradients \cite{he2016deep}
. An additional advantage of ResNets lies in their incorporation of skip connections, which helps alleviate the degradation problem associated with the worsening performance of models as complexity increases \cite{he2016deep}. Notably, unlike conventional Residual Network architectures, the Dilated ResNet incorporates dilated kernels. Dilations play a crucial role in maintaining feature maps' spatial resolution during convolutions while expanding the network's receptive field to capture more intricate details \cite{Yu2017DilatedRN}. More details of the architecture of Dilated ResNet can be found in \cite{Venkatayogi2022PitPatternCO}.  To verify our model's performance relative to other commonly used architectures for image classification tasks, we compared the performance metrics described in Sec. \ref{sec:eval} for our architecture with ResNet18 \cite{He2016DeepRL}, and AlexNet \cite{Krizhevsky2012ImageNetCW}. 

\begin{figure}[t]
		\centering
            \vspace{0.5em}
		\includegraphics[width=1\linewidth]{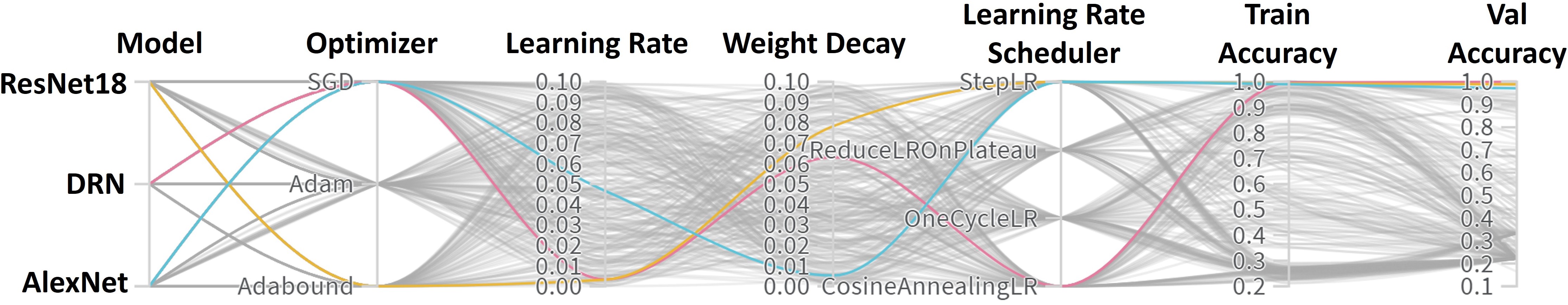}
		\vspace*{-6mm}
\caption{Hyperparameter search for the three models. Each line encodes a particular configuration.}
\label{fig:hparams}
    \vspace*{-8mm}
\end{figure}

\subsection{Evaluation}

\label{sec:eval}
To limit the variability in performance due to the variance in hyperparameters, we used a random search across our hyperparameter space to find a good candidate for each model for comparison. The hyperparameter space in this study included the initial learning rate (LR), the LR scheduler, the optimizer, and weight decay. We considered four possible LR schedulers, namely Step
, Reduce-on-Plateau, OneCycle \cite{Smith2018SuperconvergenceVF}, and Cosine Annealing \cite{Loshchilov2016SGDRSG}. Three methods for the optimizer were considered, which were Stochastic Gradient Descent (SGD), Adam \cite{Kingma2014AdamAM}, and Adabound \cite{Luo2019AdaptiveGM}. The full hyperparameter space is provided in Table \ref{tab:hparams}. Furthermore, early stopping with a patience of 10 epochs was utilized to cut down on the training time in case the validation loss reached a point of no improvement well before the designated maximum of 50 epochs. 100 random configurations for each model architecture were considered in the random search. The best configuration was chosen such that the trained model minimized the validation set loss while keeping minimal overfit (which is the difference between the training and validation accuracy). Following this step, to ensure generalizable performance, each model was configured using the chosen hyperparameters and trained using stratified 5-fold cross-validation over the training split of data such that the class distribution in each fold remained intact. After verification of generalized performance, the entire training set was used to build the final models to be used for comparative evaluation.

\begin{table}[t]
\vspace{0.5em}
\centering
\caption{Best Performing Configuration}
\begin{tabular}{@{}lccc@{}}
\toprule
\textbf{Hyperparameter} & \textbf{Dilated ResNet} & \textbf{ResNet18} & \textbf{AlexNet} \\ \midrule
LR                       & 0.06308                 & 0.07825            & 0.00527          \\
LR Scheduler             & Cosine Annealing        & Step               & Step             \\
Optimizer                & SGD                     & Adabound           & SGD              \\
Weight Decay             & 0.00338                 & 0.00324            & 0.04663          \\ \bottomrule
\end{tabular}
\label{tab:hparams_result}
\end{table}

\begin{figure}[t!]
		\centering
            \vspace*{-1em}
		\includegraphics[width=0.9\linewidth]{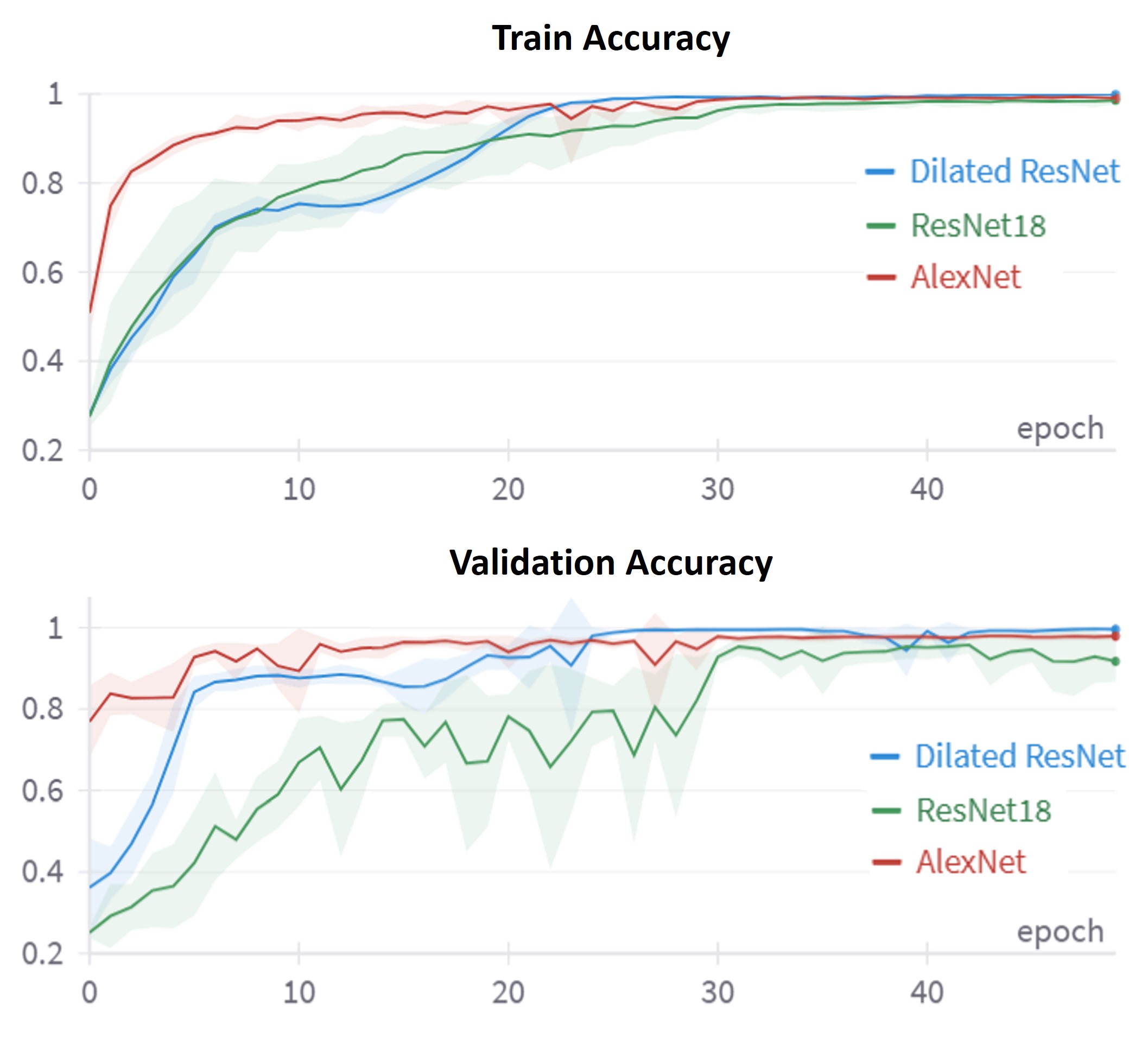}
		\vspace*{-3mm}
\caption{Stratified 5-fold cross-validation results for the three models. Average accuracy curves are reported, with the shaded region depicting the standard deviation across folds.}
\label{fig:kfold}
    \vspace*{-2em}
\end{figure}

\begin{figure}[t!]
		\centering
  		\vspace*{3mm}
		\includegraphics[width=0.9\linewidth]{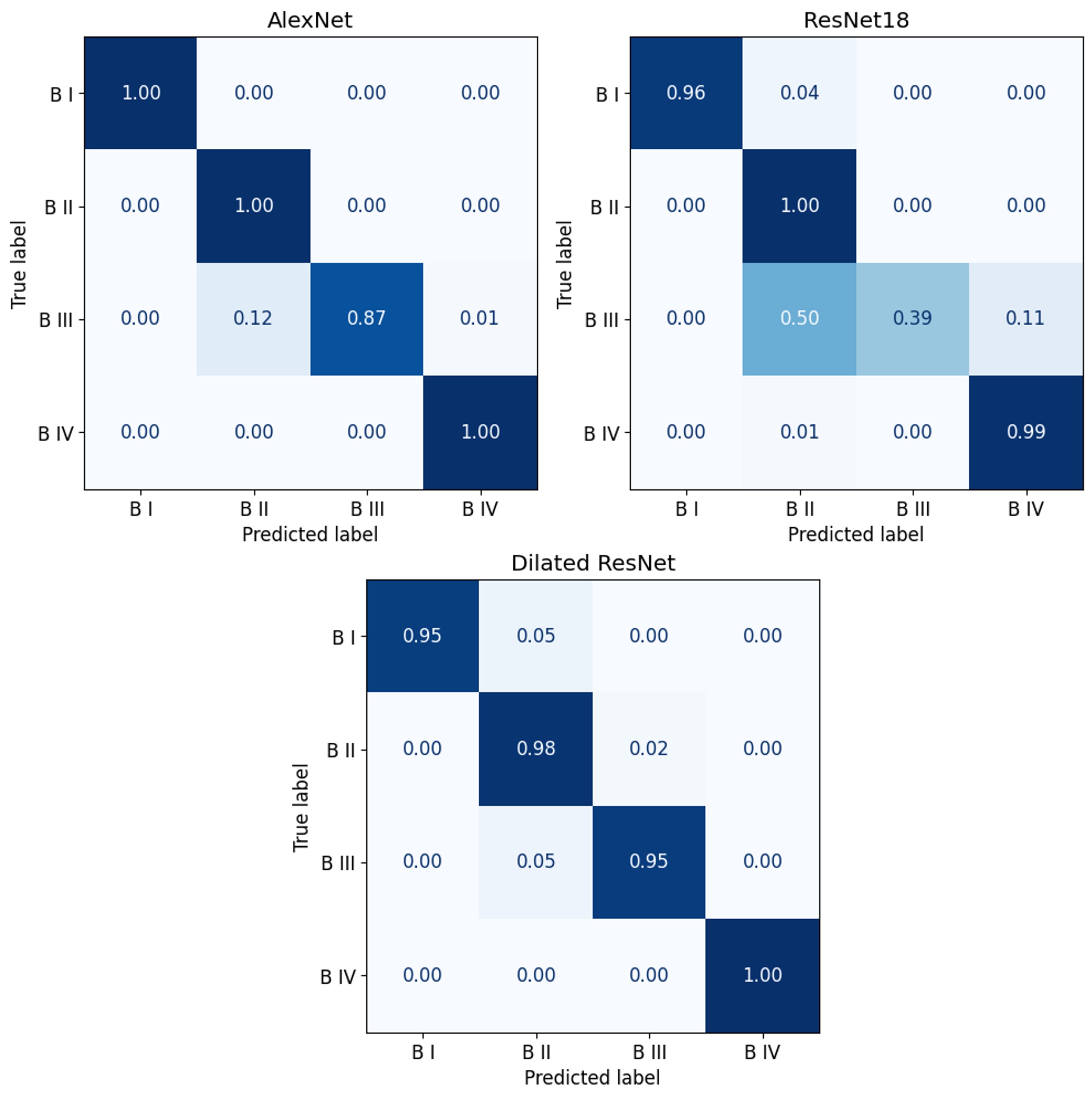}
		\vspace*{-2mm}
\caption{Normalized confusion matrices for all three models configured with their corresponding best hyperparameters. }
\label{fig:confmat}
\vspace{-2em}
\end{figure}

To evaluate model performance, we used the standard evaluation metrics of accuracy ($A$), precision ($P$), Recall ($Re$), F1-score ($F1$), and Area under the ROC Curve ($AUC$). Since this is a multi-class classification problem we reported the macro-averaged values (that is, the metrics are calculated for each class and then averaged). These metrics are crucial in assessing the model's performance in AGC tumor classification, offering a more comprehensive evaluation than accuracy alone. 
Accuracy measures overall correctness, but in safety-critical applications like cancer diagnosis, emphasizing true positives and true negatives is vital. 
Apart from these metrics, we also plotted the confusion matrices for each model to complete the analysis. The entire suite of training and tests was tracked using Weights and Biases \cite{wandb}.

\section{RESULTS AND DISCUSSION}

\subsubsection{Hyperparameter Search}
\label{sec:hparams}
Table \ref{tab:hparams_result} lists the best-performing combination of hyperparameters for each model. For the dilated ResNet, the best LR scheduler was Cosine Annealing while SGD was the best optimizer. The ResNet18 model performed best using the Step LR scheduler and Adabound optimizer. Finally, AlexNet also performed best using the Step LR scheduler coupled with an SGD optimizer. Notably, both the dilated ResNet and ResNet 18 models required an initial learning rate of around 0.06-0.07 and a weight decay of around 0.03. On the other hand, AlexNet performed better with a relatively lower learning rate of 0.005, while having a higher weight decay of 0.047.

\subsubsection{Cross Validation}
\label{sec:kfold}
The average plots of training and validation accuracy for all three candidate models training using stratified 5-fold cross-validation with the hyperparameter configuration selected as described in Sec. \ref{sec:hparams} are provided in Fig. \ref{fig:kfold}. Here we observe that ResNet18 has the largest variation in performance across the different folds, making it the most susceptible to changes in training data. The dilated ResNet model has the least variation and therefore we can conclude that it is the most general. Furthermore, the average accuracy across all folds is the lowest for ResNet18 and most for the dilated ResNet. Here again, the performance of AlexNet lies in between the other two models. While AlexNet reaches saturation much earlier than both the other models, its variation in performance across the different folds, and the lower peak accuracy as compared to the dilated ResNet18 hinder it from being the best overall pick. 
\subsubsection{Model comparisons}

The results on the test set for the three models are summarized in Table \ref{tab:results} and the corresponding confusion matrices are provided in Fig. \ref{fig:confmat}. As can be seen through our evaluations, the dilated ResNet model proposed in our previous work for colorectal cancer classification is able to outperform both ResNet18 and AlexNet for AGC tumor classification. Observing the drop in testing accuracy for ResNet18, we can conclude that it was overfitting during training. While AlexNet achieved comparable performance in terms of the performance metrics, its susceptibility to training data as discussed in Sec. \ref{sec:kfold} made it unsuitable for our application. Furthermore, the number of trainable parameters in dilated ResNet is only 2.8M compared to 11.2M and 57M trainable parameters in ResNet and AlexNet respectively. Thus, the dilated ResNet is able to outperform the other two models despite being comparatively lightweight and less complex. Looking at the confusion matrices, we observe that all three models have some difficulties in differentiating between Types II and III. This is expected since both tumor types present similarly on the exterior and there can be an overlap in diagnosis even amongst clinicians \cite{Gore2013StomachMT}.

\begin{table}[t]
\vspace{0.5em}
\centering
\caption{Performance Results}
\begin{tabular}{@{}lccc@{}}
\toprule
\textbf{Metric} & \textbf{Dilated ResNet} & \textbf{ResNet18} & \textbf{AlexNet} \\ \midrule
Accuracy         & \textbf{0.9667}          & 0.8333             & 0.9600            \\
Precision        & \textbf{0.9656}          & 0.8685             & 0.9631            \\
Recall           & \textbf{0.9692}          & 0.8175             & 0.9550            \\
F1 Score         & \textbf{0.9673}          & 0.8422             & 0.9591            \\
AUC              & \textbf{0.9990}          & 0.9879             & 0.9983            \\ \bottomrule
\end{tabular}
\label{tab:results}
\vspace{-2.5em}
\end{table}

\section{CONCLUSIONS}
To collectively address the existing limitations on endoscopic diagnosis of AGC tumors, for the first time, we used and evaluated our recently developed VTS for  classifying AGC tumors using their textural features. Leveraging a 7 DoF robotic manipulator and unique custom-designed realistic AGC tumor phantoms, we demonstrated the advantages of automated synthetic data collection using the VTS addressing the problem of data scarcity and biases encountered in traditional ML-based approaches. We also proposed and evaluated a complementary ML-based diagnostic tool that can leverage this new modality to sensitively classify AGC lesions.  Our synthetic-data-trained ML model was successfully evaluated utilizing appropriate statistical metrics even under mixed morphological characteristics and partial sensor contact. In the future, we plan to test this ML model textural images collected on real patients and perform a Sim2Real study. 

\addtolength{\textheight}{-6cm}   




\bibliographystyle{IEEEtran}
\bibliography{ISMR2023_ML}

\end{document}